\documentclass[conference]{IEEEtran}

\usepackage{cite}
\usepackage{amsmath,amssymb,amsfonts}
\usepackage{algorithmic}
\usepackage{graphicx}
\usepackage{textcomp}
\usepackage{xcolor}
\def\BibTeX{{\rm B\kern-.05em{\sc i\kern-.025em b}\kern-.08em
    T\kern-.1667em\lower.7ex\hbox{E}\kern-.125emX}}

\usepackage{microtype}
\usepackage{graphicx}
\usepackage{subfigure}
\usepackage{booktabs} 
\usepackage{enumitem}

\usepackage{hyperref}

\usepackage{amsmath,amsfonts,bm}

\def\eqref#1{equation~\ref{#1}}

\def\1{\bm{1}}

\def\rvy{{\mathbf{y}}}

\def\ervy{{\textnormal{y}}}

\def\rmE{{\mathbf{E}}}

\def\ermE{{\textnormal{E}}}

\def\vc{{\bm{c}}}

\def\ve{{\bm{e}}}

\def\evc{{c}}

\def\mM{{\bm{M}}}

\def\mQ{{\bm{Q}}}

\def\mS{{\bm{S}}}

\DeclareMathAlphabet{\mathsfit}{\encodingdefault}{\sfdefault}{m}{sl}
\SetMathAlphabet{\mathsfit}{bold}{\encodingdefault}{\sfdefault}{bx}{n}

\def\emM{{M}}

\def\emQ{{Q}}

\def\emS{{S}}

\newcommand{\KL}{D_{\mathrm{KL}}}

\DeclareMathOperator*{\argmin}{arg\,min}

\usepackage{amsmath}
\usepackage{amssymb}
\usepackage{mathtools}
\usepackage{amsthm}
\usepackage{thmtools}
\usepackage{thm-restate}

\usepackage{algorithm}

\usepackage[capitalize,noabbrev]{cleveref}

\theoremstyle{plain}

\theoremstyle{definition}

\theoremstyle{remark}

\definecolor{mydarkblue}{rgb}{0,0.08,0.45}
\hypersetup{ %
pdftitle={},
pdfsubject={},
pdfkeywords={},
pdfborder=0 0 0,
pdfpagemode=UseNone,
colorlinks=true,
linkcolor=mydarkblue,
citecolor=mydarkblue,
filecolor=mydarkblue,
urlcolor=mydarkblue,
}

\begin{document}
\bstctlcite{NoDashCtl}

\title{Information-Theoretic Active Correlation Clustering}

\author{
\IEEEauthorblockN{Linus Aronsson}
\IEEEauthorblockA{Department of Computer Science and Engineering\\
Chalmers University of Technology \& \\
University of Gothenburg\\
Gothenburg, Sweden\\
linaro@chalmers.se}
\and
\IEEEauthorblockN{Morteza Haghir Chehreghani}
\IEEEauthorblockA{Department of Computer Science and Engineering\\
Chalmers University of Technology \& \\
University of Gothenburg\\
Gothenburg, Sweden\\
morteza.chehreghani@chalmers.se}
}


\maketitle
\IEEEpeerreviewmaketitle 

\begin{abstract}
Correlation clustering is a flexible framework for partitioning data based solely on pairwise similarity or dissimilarity information, without requiring the number of clusters as input. However, in many practical scenarios, these pairwise similarities are not available a priori and must be obtained through costly measurements or human feedback. This motivates the use of active learning to query only the most informative pairwise comparisons, enabling effective clustering under budget constraints. In this work, we develop a principled active learning approach for correlation clustering by introducing several information-theoretic acquisition functions that prioritize queries based on entropy and expected information gain. These strategies aim to reduce uncertainty about the clustering structure as efficiently as possible. We evaluate our methods across a range of synthetic and real-world settings and show that they significantly outperform existing baselines in terms of clustering accuracy and query efficiency. Our results highlight the benefits of combining active learning with correlation clustering in settings where similarity information is costly or limited.
\end{abstract}


\section{Introduction} \label{section:introduction}

Clustering is an important unsupervised learning problem for which several methods have been proposed in different contexts.
\emph{Correlation clustering} (CC) \cite{BansalBC04,DemaineEFI06} is a well-known clustering problem, especially beneficial when both similarity and dissimilarity assessments exist for a given set of $N$ objects. Consequently, CC studies the clustering of objects where pairwise similarities can manifest as positive or negative numbers. 
It has found a wide range of applications including image segmentation \cite{KimNKY11}, bioinformatics \cite{BonchiGU13}, spam filtering \cite{BonchiGL14}, social network analysis \cite{2339530.2339735, 2956185}, duplicate detection \cite{HassanzadehCML09}, co-reference identification \cite{McCallumW04}, entity resolution \cite{GetoorM12}, color naming across languages \cite{ThielCD19} and clustering aggregation \cite{GionisMT07,ChehreghaniC20}. 
CC was initially explored using binary pairwise similarities in $\{-1, +1\}$ \cite{BansalBC04}, and was later extended to support arbitrary positive and negative pairwise similarities in $\mathbb{R}$ \cite{CharikarGW05, DemaineEFI06}. Finding the optimal solution for CC is known to be NP-hard and APX-hard \cite{BansalBC04,DemaineEFI06}, presenting significant challenges. As a result, various approximate algorithms have been developed to address this problem \cite{BansalBC04, CharikarGW05,DemaineEFI06,AilonCN08,elsner-schudy-2009-bounding}. Among these, methods based on local search are noted for their superior performance in terms of clustering quality and computational efficiency \cite{ThielCD19,Chehreghani22_shift}.

Existing methods generally assume that all $N\choose2$ pairwise similarities are available beforehand. However, as discussed in \cite{bressan2020,bonchi2020}, generating pairwise similarities can be computationally intensive and may need to be obtained through resource-intensive queries, e.g., from a human expert. For instance, determining interactions between biological entities often requires the expertise of highly trained professionals, consuming both time and valuable resources \cite{bonchi2020}. In tasks like entity resolution, obtaining pairwise similarity queries through crowd-sourcing could also involve monetary costs. Therefore, a central question emerges: \emph{How can we design a machine learning paradigm that effectively delivers satisfactory CC results with a limited number of queries for pairwise similarities between objects?}

In machine learning, \emph{active learning} is generally employed to address such a question. Its objective is to acquire the most informative data within a constrained budget. 
Active learning has proven effective in various tasks, including recommender systems \cite{Rubens2015}, sound event detection \cite{TASLP.2020.3029652}, analysis of driving time series \cite{JarlARC22}, drug discovery \cite{minf.202200043}, and analysis of logged data  \cite{YanCJ18}. In the context of active learning, the selection of which data to query is guided by an \emph{acquisition function}. Active learning is most commonly studied for classification and regression problems \cite{settles.tr09}. However, it has also been studied for clustering and is sometimes referred to as \emph{supervised clustering} \cite{AwasthiZ10}. The objective is to discover the ground-truth clustering with a minimal number of queries to an \emph{oracle} (e.g., a human expert). 
In this scenario, queries are typically executed in one of two ways: (i) By asking whether two clusters should merge or if one cluster should be divided into multiple clusters \cite{balcan2008, AwasthiZ10, awasthi2017}; (ii) By querying the pairwise relations between objects \cite{Basu2004ActiveSF, mazumdar2017-2, mazumdar2017-1, saha2019, bressan2020, bonchi2020, Craenendonck2018COBRASIC, silwal2023kwikbucks, mabcc, anonymous, noisyqecc}.

Among the aforementioned works on active learning for clustering, only \cite{mazumdar2017-2, bressan2020, bonchi2020, anonymous, noisyqecc} consider the setting that we are interested in: (i) The clustering algorithm is based on CC; (ii) The pairwise similarities are not assumed to be known in advance; (iii) We assume access to a single noisy oracle, to which a \emph{fixed} budget $B \ll $ $N\choose2$ of queries for pairwise similarities can be performed; (iv) Access to feature vectors is not assumed by the algorithm, meaning that information about the ground-truth clustering is solely obtained through querying the oracle for pairwise similarities. Throughout the paper, this setting will be referred to as \emph{active correlation clustering}. 

The work in \cite{mazumdar2017-2} develops a number of \emph{pivot-based} CC algorithms that satisfy guarantees on the query complexity, assuming a noisy oracle. However, the algorithms are purely theoretical and are not implemented and investigated in practice, and require setting a number of non-trivial parameters (e.g., they assume the noise level is known in advance which is unrealistic). The work in \cite{bressan2020, bonchi2020} proposes adaptive and query-efficient versions of the simple pivot-based CC algorithm KwikCluster \cite{AilonCN08}. However, as demonstrated in \cite{anonymous}, such pivot-based methods perform very poorly for active CC with noise. The work in \cite{mabcc, noisyqecc} address query-efficient CC by formulating it as a \textit{multi-armed bandit} problem. However, this leads to a number of limiting assumptions in practice, as the goal is to provide theoretical guarantees.

The work in \cite{anonymous} proposes a generic active CC framework that overcomes the limitations of previous work and offers several advantages: (i) The pairwise similarities can be any positive or negative real number, even allowing for inconsistencies (i.e., violation of transitivity). This allows the oracle to express uncertainty in their feedback; (ii) The process of querying pairwise similarities is decoupled from the clustering algorithm, enhancing flexibility in constructing acquisition functions that can be employed in conjunction with \emph{any} CC algorithm. \cite{anonymous} employs an efficient CC algorithm based on local search, whose effectiveness (and superiority over pivot-based methods) has also been demonstrated in the standard CC setting \cite{ThielCD19,Chehreghani22_shift}, and dynamically computes the number of clusters; (iii) The framework is robust w.r.t. a noisy oracle and supports multiple queries for the same pairwise similarity if needed (to deal with noise).

Furthermore, \cite{anonymous} proposes two novel acquisition functions, namely \emph{maxmin} and \emph{maxexp}, to be used within their framework. They demonstrate that the algorithm QECC from \cite{bonchi2020} performs poorly in the presence of even a very small amount of noise and is significantly outperformed by their methods. 
In this paper, we adopt the generic active CC framework in \cite{anonymous} with a focus on the development of more effective acquisition functions. The contributions of this paper are the following:

\begin{itemize}[leftmargin=*]
\item We investigate the use of information-theoretic acquisition functions based on \emph{entropy} and \emph{information gain} for active CC. We propose three different acquisition functions inspired by this (see Section \ref{section:information-theoretic}). Although information-theoretic acquisition functions have been extensively studied in the context of active learning \cite{mccallum, kirsch2022unifying}, prior research has focused mainly on (active) \emph{supervised learning} scenarios, where the goal is to query data labels from an oracle rather than pairwise relations. \emph{To our knowledge, our work is the first attempt to propose information-theoretic acquisition functions to active learning with pairwise relations, as well as to non-parametric models like CC.} Computing the necessary quantities in this setting is significantly more complex. The methods proposed in this paper can be applied beyond active CC, including to the active learning of other pairwise (non-parametric) clustering models. 

\item We conduct extensive experimental studies on various datasets that demonstrate the superior performance of our acquisition functions compared to \emph{maxmin} and \emph{maxexp} (and other baselines), and investigate a number of interesting insights about the active CC framework from \cite{anonymous}.
\end{itemize}

\section{Active Correlation Clustering} \label{section:acc-form}

In this section, we begin by introducing the problem of active CC. After this, we describe the active clustering procedure used to solve this problem.

\subsection{Problem Formulation}

We are given a set of $N$ objects (data points) indexed by $\mathcal{V} = \{1,\hdots,N\}$. The set of pairs of objects in $\mathcal{V}$ is denoted by $\mathcal{E}=\{(u,v) \; | \; u,v\in \mathcal{V}\}$. We assume the existence of a ground-truth similarity matrix $\mS^{\ast} \in \mathbb{R}^{N \times N}$, which represents the true pairwise similarities between every pair $(u, v) \in \mathcal{E}$. However, $\mS^{\ast}$ is not known beforehand. Instead, one can only query the oracle for a noisy version of this matrix for a desired pair of objects, while incurring some cost. We use $\mS \in \mathbb{R}^{N \times N}$ to represent an estimate of the pairwise similarities. If $\emS_{uv} = \emS^{\ast}_{uv}$ for all $(u,v) \in \mathcal{E}$ we have a perfect estimate of the true pairwise similarities, which we assume is unrealistic in practice. Hence, the objective is to discover the ground-truth clustering solution with a minimal number of (active) queries for the pairwise similarities to the oracle, since each query incurs some cost. A similarity matrix $\mS$ is symmetric, and we assume zeros on the diagonal, i.e., $\emS_{uv} = \emS_{vu}$ and $\emS_{uu} = 0$. This means there are ${N \choose 2} = (N \times (N-1))/2$ unique pairwise similarities to estimate. Without loss of generality, we assume all similarities are in the range $[-1, +1]$. In this case, $+1$ and $-1$ respectively indicate definite similarity and dissimilarity. Thus, a similarity close to $0$ indicates a lack of knowledge about the relation between the two objects. This allows the oracle to express uncertainty in their feedback.

A clustering is a partition of $\mathcal{V}$. In this paper, we encode a clustering with $K$ clusters as a clustering solution $\vc \in [K]^{N}$ where $[K] = \{1,\hdots,K\}$ and $c_u \in [K]$ denotes the cluster label of object $u \in \mathcal{V}$. We denote by $\mathcal{C}$ the set of clustering solutions for all possible partitions (clusterings) of $\mathcal{V}$. We say a pair $(u, v)$ violates a clustering $\vc$ if $c_u = c_v \text{ and } S_{uv} < 0 \text{ or } c_u \neq c_v \text{ and } S_{uv} \geq 0$. Given a clustering solution \(\vc \in \mathcal{C}\), the CC cost function $R^{\text{CC}}: \mathcal{C} \rightarrow \mathbb{R}^+$ aims to penalize cluster disagreements, as shown in Eq. \ref{eq:cost}.

\begin{equation} \label{eq:cost}
R^{\text{CC}}(\boldsymbol{c} \mid \boldsymbol{S}) \triangleq \sum_{(u, v) \in \mathcal{E}} \begin{cases} 
|S_{uv}| & \text{if } (u, v) \text{ violates } \vc \\
0 & \text{otherwise.}
\end{cases}
\end{equation}

\begin{restatable}{proposition}{propmaxcorr} \label{prop:maxcorr} Eq. \ref{eq:cost} can be simplified to $R^{\text{CC}}(\vc \mid \mS)= -\sum_{\substack{(u, v) \in \mathcal{E}\\c_u=c_v}} \emS_{uv} + \text{constant}$, where the constant is independent of different clustering solutions \cite{ethz-a-010077098,Chehreghani22_shift}.
\end{restatable}

Based on Proposition \ref{prop:maxcorr}\footnote{All proofs can be found in Appendix \ref{appendix:proofs}.}, we define the \textit{max correlation} cost function as 
\begin{equation} \label{eq:maxcorrfn}
R^{\text{MC}}(\vc \mid \mS) \triangleq -\sum_{\substack{(u, v) \in \mathcal{E}\\c_u=c_v}} \emS_{uv},
\end{equation}
and we have $\operatorname*{argmin}_{\vc \in \mathcal{C}} \; R^{\text{CC}}(\vc \mid \mS) = \operatorname*{argmin}_{\vc \in \mathcal{C}} \;  R^{\text{MC}}(\vc \mid \mS)$. Because of this, we will use $R^{\text{MC}}$ throughout most of the paper, as it leads to a number of simplifications in the presented methods. The conditioning on $\mS$ for $R^{\text{CC}}$ and $R^{\text{MC}}$ will often be dropped, unless it is not clear from context. Finally, the ground-truth clustering solution corresponds to $\vc^{\ast} =  \operatorname*{argmin}_{\vc \in \mathcal{C}} \;  R^{\text{MC}}(\vc \mid \mS^{\ast})$.

\subsection{Active Correlation Clustering Procedure} \label{section:acc-procedure}

\begin{algorithm}[tb]
\caption{Active CC} \label{alg:acc}
\begin{algorithmic}[1]
\STATE {\bfseries Input:} Initial similarity matrix $\mS^0$, acquisition function $a$, batch size $B$.
\STATE $i \coloneqq 0$
\WHILE{query budget not reached}
    \STATE $\vc^i \coloneqq $ \hyperref[alg:clustering-alg]{CC}$(\mS^i)$ \COMMENT{Alg. \ref{alg:clustering-alg}}
    \STATE $\mathcal{B} \coloneqq \operatorname*{argmax}_{\mathcal{B} \subseteq \mathcal{E}, |\mathcal{B}| = B} \sum_{(u, v) \in \mathcal{B}} a(u, v)$
    \STATE Query (noisy) oracle and update $\emS_{uv}^{i+1}$ for all pairs $(u, v) \in \mathcal{B}$
    \STATE $i \coloneqq i + 1$
\ENDWHILE \\
\textbf{Return} $\vc^i$
\end{algorithmic}
\end{algorithm}

\begin{algorithm}[tb]
   \caption{Local Search for CC}
   \label{alg:clustering-alg}
\begin{algorithmic}[1]
    \STATE Randomly assign each object $i \in \mathcal{V}$ to a cluster.
    \WHILE{not converged}
        \STATE Select object $i \in \mathcal{V}$ randomly
        \STATE Assign object $i$ to the cluster that maximally increases the objective in Eq. \ref{eq:maxcorrfn}. If no existing cluster yields an increase in the objective, form a new cluster with $i$ as its only member. \COMMENT{In practice, this step is implemented as in \cite{anonymous}: by considering only the terms in Eq.~\ref{eq:maxcorrfn} that involve object \( i \), the computational complexity of each iteration is reduced from \( O(K^2N^2) \) to \( O(KN) \).}
    \ENDWHILE
\end{algorithmic}
\end{algorithm}

We adopt the recent generic active CC procedure outlined in \cite{anonymous} to solve the problem described in the previous section. The procedure is shown in Alg. \ref{alg:acc}. It takes an initial similarity matrix $\mS^0$ as input, which can contain partial or no information about $\mS^{\ast}$, depending on the initialization method. The procedure then follows a number of iterations, where each iteration $i$ consists of three steps: (i) \textit{Update} the current clustering solution $\vc^i \in \mathcal{C}$ by running a CC algorithm given the current similarity matrix $\mS^i$. The current similarity matrix $\mS^i$ will be referred to as $\mS$ throughout the paper; (ii) \textit{Select} a batch $\mathcal{B} \subseteq \mathcal{E}$ of pairs of size $B = |\mathcal{B}|$ based on an acquisition function $a : \mathcal{E} \rightarrow \mathbb{R}$. The quantity $a(u, v)$ indicates how informative the pair $(u, v) \in \mathcal{E}$ is, where a higher value implies greater informativeness. The optimal batch is selected by $\mathcal{B} = \operatorname*{argmax}_{\mathcal{B} \subseteq \mathcal{E}, |\mathcal{B}| = B} \sum_{(u, v) \in \mathcal{B}} a(u, v)$. This corresponds to selecting the top-$B$ pairs based on their acquisition value; (iii) \textit{Query} the oracle for the pairwise similarities of the pairs $(u, v) \in \mathcal{B}$ and update each $\emS_{uv}^{i+1}$ based on the response.

\section{Information-Theoretic Acquisition Functions}
\label{section:information-theoretic}

In this section, we introduce three information-theoretic acquisition functions for active CC. All quantities defined below are conditioned on the current similarity matrix $\mS$, but it is left out for brevity. All acquisition functions proposed in this section depend on the Gibbs distribution defined as
\begin{align} \label{eq:gibbs}
    P^{\text{Gibbs}}(\rvy = \vc) \triangleq \frac{\exp(-\beta R^{\text{MC}}(\vc))} {\sum_{\vc^{\prime} \in \mathcal{C}} \exp(-\beta R^{\text{MC}}(\vc^{\prime}))},
\end{align}
where $\beta \in \mathbb{R}^+$ is a concentration parameter, $\rvy = \{\ervy_1,\hdots,\ervy_N\}$ is a random vector with sample space $\mathcal{C}$ (all possible clustering solutions of $\mathcal{V}$) and $\ervy_u$ is a random variable for the cluster label of $u$ with sample space $[K]$. Computing $P^{\text{Gibbs}}$ is intractable due to the sum over all possible clustering solutions $\mathcal{C}$ in the denominator. Therefore, in the next section, we describe a \emph{mean-field approximation} of $P^{\text{Gibbs}}$ which makes it possible to efficiently calculate the proposed acquisition functions. To the best of our knowledge, the use of mean-field approximation to approximate complex quantities when applying information-theoretic acquisition functions for active learning is a novel aspect of our approach. This approach
can be applied beyond active CC, extending to the active learning of other pairwise, non-parametric clustering models.

\subsection{Mean-Field Approximation for CC}
\label{sec:mean_field}


We here describe the mean-field approximation of $P^{\text{Gibbs}}$. The family of factorial distributions over the space of clustering solutions is defined as 
\begin{equation}
\mathcal{Q}=\{Q \in \mathcal{P} \mid Q(\rvy = \vc)=\prod_{u \in \mathcal{V}} Q(\ervy_u = \evc_u)\},
\end{equation}
where $\mathcal{P}$ is the space of all probability distributions with sample space $\mathcal{C}$. The goal of mean-field approximation is to find a factorial distribution $Q \in \mathcal{Q}$ that best approximates the intractable distribution $P^{\text{Gibbs}}$. In general, one can compute the optimal $Q$ by minimizing the KL-divergence \cite{hofman, pmlr-v22-haghir12}, i.e., 
\begin{equation} \label{eq:kl}
Q^* = \argmin_{Q \in \mathcal{Q}} \KL(Q\| P^{\text{Gibbs}}).
\end{equation}
%
We encode a mean-field approximation using a matrix of assignment probabilities $\mQ \in [0, 1]^{N \times K}$, where $Q_{uk} = Q(\ervy_u = k)$. In addition, let $\mM \in \mathbb{R}^{N \times K}$, where $M_{uk}$ should be interpreted as the cost of assigning object $u$ to cluster $k$. Given this, Theorem \ref{prop:kl} implies that an EM-type procedure,  which sequentially alternates between estimating $Q_{uk}$ (based on Eq. \ref{eq:EM1}) and computing the respective $M_{uk}$ (based on Eq. \ref{eq:EM2}), yields a local minimum
for the optimization problem in Eq. \ref{eq:kl}. In Theorem \ref{prop:kl}, we adapt and specialize the general result from \cite{DBLP:journals/pami/HofmannPB98} to our specific cost function in Eq. \ref{eq:maxcorrfn}, enabling efficient mean-field approximations tailored to our model, which are essential for all proposed acquisition functions.


\begin{restatable}{theorem}{propkl} \label{prop:kl}
\label{thm:async_convergence}
Let $\ell : \mathbb{N} \rightarrow \mathcal{V}$ denote an object visitation schedule, which satisfies $\lim_{T \rightarrow \infty}  |\{ t \leq T : \ell(t) = u \}| = \infty, \forall u \in \mathcal{V}$. For arbitrary initial conditions, the asynchronous update rules defined by
\begin{align}
    Q^{(t+1)}_{uk} &=\exp(-\beta M^{(t)}_{uk})/\sum_{k^{\prime} \in [K]} \exp(-\beta M^{(t)}_{uk^{\prime}}), \label{eq:EM1}\\
    M^{(t+1)}_{uk} &= -\sum_{\substack{v \in \mathcal{V} \\ v \neq u}} S_{uv} Q_{vk}^{(t+1)}, \label{eq:EM2}
\end{align}
where $u = \ell(t)$, converge to a local minimum of Eq. \ref{eq:kl}.
\end{restatable}
%
\begin{algorithm}[tb]
\caption{Mean-Field Approximation} \label{alg:mf}
\begin{algorithmic}[1]
\STATE {\bfseries Input:} Similarity matrix $\mS$, cluster assignment costs $\mM$, concentration parameter $\beta$.
\WHILE{$\mQ$ has not converged}
    \STATE $\mQ \coloneqq \text{softmax}(-\beta\mM)$\COMMENT{E-step}
    \STATE $\mM \coloneqq -\mS \cdot \mQ$ \COMMENT{M-step}
\ENDWHILE \\
\textbf{Return} $\mQ, \mM$
\end{algorithmic}
\end{algorithm}
%
Note that we use a different formulation of correlation clustering in the mean-field approximation than the one used in \cite{pmlr-v22-haghir12}, which results in improved computational efficiency. To further enhance efficiency, we employ a synchronous update rule in practice (see Alg. \ref{alg:mf}). Despite not having the same theoretical guarantees, synchronous updates have been observed to perform well empirically in other contexts \cite{DBLP:journals/pami/HofmannPB98,pmlr-v22-haghir12}. Alg. \ref{alg:mf} assumes a fixed number of clusters $K$. We use the number of clusters $K$ dynamically determined by the CC algorithm used at each iteration $i$ of Alg. \ref{alg:acc} to find $\vc^i$ (Alg. \ref{alg:clustering-alg}). $\mM$ could be initialized randomly. However, since we have the current clustering solution $\vc^i$, we initialize it based on $\vc^i$, i.e., $\emM_{uk} = -\sum_{v : c^i_v = k} \emS_{uv}$, in order to speed up the convergence and potentially improve the quality of the solution found. This initialization of $\mM$ is based on the total similarity between object $u$ and cluster $k$ in relation to the similarity between $u$ and all other clusters. A smaller similarity should correspond to a higher cost (hence the negation). Each iteration of the algorithm consists of two main steps. First, $\mQ$ is estimated as a function of $\mM$. Second, $\mM$ is calculated based on $\mQ$. In this paper, we treat the concentration parameter $\beta \in \mathbb{R}^+$ as a hyperparameter. Finally, we employ the special form of the max correlation cost function $ R^{\text{MC}}$ in Eq. \ref{eq:maxcorrfn}, and calculate both the E-step and M-step in vectorized form. In particular, the M-step becomes a dot product between $\mS$ and $\mQ$, which is extremely efficient in practice (especially if $\mS$ is assumed sparse, which it is in our experiments). 

\subsection{Entropy} \label{section:entropy}

In this section, we propose our first acquisition function based on entropy. Let $\rmE \in \{-1, +1\}^{N \times N}$ be a random matrix where each element $\ermE_{uv}  \in \{-1, +1\}$ is a binary random variable, where $+1$ indicates $u$ and $v$ should be in the same cluster, and $-1$ implies $u$ and $v$ should be in different clusters. A reasonable way to define the probability of $\ermE_{uv}$ to be $+1$ is the fraction of clustering solutions in $\mathcal{C}$ that assign $u$ and $v$ to the same cluster, weighted by the probability of each clustering solution. Due to the intractability of  $P^{\text{Gibbs}}$, we approximate it using a mean-field approximation $Q$ (encoded by matrix $\mQ$, as described in the previous section). Formally, we have
\begin{equation} \label{eq:peuv}
\begin{aligned}
    P(E_{uv} = 1) &= \mathbb{E}_{P^{\text{Gibbs}}(\rvy)}[\1_{\{\ervy_u = \ervy_v\}}] \\
    &\approx \mathbb{E}_{Q(\rvy)}[\1_{\{\ervy_u = \ervy_v\}}] \\
    &= \sum_{k \in [K]} Q_{uk}Q_{vk}.
\end{aligned}
\end{equation}
The last equality of Eq. \ref{eq:peuv} uses the fact that the mean-field approximation assumes independence between objects. One can similarly derive $P(\ermE_{uv} = -1) = \sum_{k,k^{\prime} \in [K]} Q_{uk}Q_{vk^{\prime}}\1_{\{k \neq k^{\prime}\}} = 1 - P(\ermE_{uv} = 1)$. Thereby, from Eq. \ref{eq:peuv}, we define an acquisition function based on the entropy of $\ermE_{uv}$ as
\begin{equation} \label{eq:entropy}
\begin{aligned}
    a^{\text{Entropy}}(u, v) &\triangleq H(\ermE_{uv})
    = \mathbb{E}_{P(\ermE_{uv})}[-\log P(\ermE_{uv})].
\end{aligned}
\end{equation}
\subsection{Information Gain} \label{section:ig}
The acquisition function $a^{\text{Entropy}}$ calculates the uncertainty of pairs based on the mean-field approximation (model) $\mQ$ given the current similarity matrix $\mS$. In this section, we investigate acquisition functions inspired by the notion of \emph{information gain} corresponding to maximal uncertainty reduction. In this case, the similarity matrix $\mS$ is first augmented with $\emph{pseudo-similarities}$ (predicted using the current model $\mQ$ as $\emS_{uv} \sim P(\ermE_{uv})$), after which a new mean-field approximation is obtained. In other words, we simulate the effect of querying one or more pairs in expectation w.r.t. the current model $\mQ$, potentially resulting in more accurate uncertainty estimations. Due to the efficiency of Alg. \ref{alg:mf} (mean-field), one can afford to run it several times per iteration of the active CC procedure, to estimate the information gain accurately. In this paper, we consider two types of information gain. First, the information gain (or equivalently the mutual information) between a pair $\ermE_{uv}$ and the cluster labels of objects $\rvy$. Due to symmetry of the mutual information we have
\begin{align}
I(\rvy;\ermE_{uv}) &= H(\rvy) - H(\rvy \mid \ermE_{uv}) \label{eq:igo1} \\
&= H(\ermE_{uv}) - H(\ermE_{uv} \mid \rvy). \label{eq:igo2}
\end{align}
The interpretation of $I(\rvy;\ermE_{uv})$ is the amount of information one expects to gain about the cluster labels of objects by observing $\ermE_{uv}$, where the expectation is w.r.t. $P(\ermE_{uv})$. In other words, it measures the expected reduction in uncertainty (in entropic way) over the possible clustering solutions w.r.t. the value of $\ermE_{uv}$. Second, the information gain between a pair $\ermE_{uv}$ and all pairs $\rmE$:
\begin{align}
I(\rmE;\ermE_{uv}) &= H(\rmE) - H(\rmE \mid \ermE_{uv}) \label{eq:JEIG1} \\
&= H(\ermE_{uv}) - H(\ermE_{uv} \mid \rmE). \label{eq:JEIG2}
\end{align}
Intuitively, $I(\rmE;\ermE_{uv})$ measures the amount of information the pair $\ermE_{uv}$ provides about all pairs in $\rmE$. All the expressions above are closely related, but the formulations used will impact how they can be approximated in practice, leading to differences in performance and efficiency. This is discussed in detail in the following subsections.

\subsubsection{Conditional Mean-Field Approximation}

We approximate all the conditional entropies defined above following the same general principle: We update the similarity matrix $\mS$ based on what is being conditioned on, run Alg. \ref{alg:mf} given this similarity matrix, and calculate the corresponding entropy given the updated mean-field approximation. Motivated by this, the following notation will be used throughout this section. Let $\ve$ denote a vector in $\{-1, +1\}^{|\mathcal{D}|}$, where $\mathcal{D} \subseteq \mathcal{E}$ is a subset of the pairs. Given this, we denote by $\mQ^{(\mS_{\mathcal{D}} = \ve)}$ to be the mean-field approximation found by Alg. \ref{alg:mf} after modifying $\mS$ according to $\ve$ for all pairs $(u, v) \in \mathcal{D}$ (with remaining pairs unchanged).

\subsubsection{Expected Information Gain} \label{section:IGO}

In this section, we focus on the expression in Eq.~\ref{eq:igo1}, which represents the \emph{expected information gain over the cluster labels of objects} (EIG-O). We have that

\begin{equation}
    H(\rvy \mid \ermE_{uv}) = \mathbb{E}_{e \sim P(\ermE_{uv})}[ H(\rvy \mid \ermE_{uv} = e)].
\end{equation}

In our framework, we approximate $H(\rvy \mid \ermE_{uv} = e)$ using the conditional mean-field approximation $\mQ^{(S_{uv} = e)}$. Given any mean-field approximation $\mQ^{\prime}$, we compute the probability of $\ermE_{uv}$ as described in Eq.~\ref{eq:peuv} using $\mQ^{\prime}$, and we define the entropy for any object $w$ as

\begin{equation}
    H(\ervy_w \mid \mQ^{\prime}) \triangleq -\sum_{k \in [K]} Q^{\prime}_{wk}\log Q^{\prime}_{wk}.
\end{equation}

Since each $\ervy_u \in \rvy$ is conditionally independent given the mean-field approximation, the joint entropy decomposes into a sum of individual entropies. That is,
\begin{equation}
    H(\rvy) \approx \sum_{w \in \mathcal{V}} H(\ervy_w \mid \mQ)
\end{equation}
and
\begin{equation}
    H(\rvy \mid \ermE_{uv} = e) \approx \sum_{w \in \mathcal{V}} H(\ervy_w \mid \mQ^{(S_{uv} = e)}).
\end{equation}
Based on this approximation, we define the following acquisition function:

\begin{equation} \label{eq:eigo}
\begin{aligned}
    &a^{\text{EIG-O}}(u, v) \triangleq  \sum_{w \in \mathcal{V}} H(\ervy_w \mid \mQ) \\
    &- \sum_{e \in \{-1, +1\}} P(\ermE_{uv} = e \mid \mQ) H(\ervy_w \mid \mQ^{(S_{uv} = e)}).
\end{aligned}
\end{equation}

The entropy of $\ermE_{wl}$ is calculated as in Eq. \ref{eq:entropy} given some mean-field approximation. 

Calculating $a^{\text{EIG-O}}$ for every pair requires running Alg.~\ref{alg:mf} $\binom{N}{2}$ times, which becomes computationally prohibitive for large $N$. In Alg.~\ref{alg:ig}, we describe practical strategies to improve this efficiency:
\begin{enumerate}[leftmargin=*, label=(\roman*)]
    \item We evaluate Eq.~\ref{eq:eigo} only for a subset of pairs $\mathcal{E}^{\text{EIG}} \subseteq \mathcal{E}$. In practice, this subset is chosen as the top-$|\mathcal{E}^{\text{EIG}}|$ pairs according to $a^{\text{Entropy}}$, with $|\mathcal{E}^{\text{EIG}}| = O(N)$.
    \item We observe that $\mQ^{(S_{uv} = e)}$ is typically similar to $\mQ$. Thus, by initializing $\mM$ (as in lines 7--8) with the assignment costs computed in line 3, the convergence rate of Alg. ~\ref{alg:mf} is significantly enhanced.
\end{enumerate}

\begin{algorithm}[t!]
\caption{EIG-O} \label{alg:ig}
\begin{algorithmic}[1]
\STATE {\bfseries Input:} Similarity matrix $\mS$, current clustering $\vc^i$, concentration parameter $\beta$.
\STATE $\emM_{uk} \coloneqq -\sum_{v : c^i_v = k} \emS_{uv}, \forall u \in \mathcal{V}, \forall k \in [K]$
\STATE $\mQ, \mM \coloneqq$ \hyperref[alg:mf]{MeanField}($\mS,\mM, \beta$)
\STATE $\mathcal{E}^{\text{EIG}} \coloneqq$ select top-$|\mathcal{E}^{\text{EIG}}|$ pairs using $a^{\text{Entropy}}$ given $\mQ$ (Eq. \ref{eq:entropy}).
\STATE $a^{\text{EIG}}(u, v) \coloneqq 0, \forall (u, v) \in \mathcal{E}$
\FOR{each pair $(u, v) \in \mathcal{E}^{\text{EIG}}$}
    \STATE $\mQ^{(\emS_{uv} = +1)} \coloneqq$ \hyperref[alg:mf]{MeanField}$(\mS, \mM, \beta \mid \emS_{uv} = +1)$
    \STATE $\mQ^{(\emS_{uv} = -1)} \coloneqq$ \hyperref[alg:mf]{MeanField}$(\mS, \mM, \beta \mid \emS_{uv} = -1)$
    \STATE $a^{\text{EIG}}(u, v) \coloneqq $ Evaluate Eq. \ref{eq:eigo} given $\mQ$, $\mQ^{(\emS_{uv} = +1)}$ and $\mQ^{(\emS_{uv} = -1)}$
\ENDFOR \\
\textbf{Return} $a^{\text{EIG}}$
\end{algorithmic}
\end{algorithm}

\begin{algorithm}[t!]
\caption{JEIG} \label{alg:JEIG}
\begin{algorithmic}[1]
\STATE \textbf{Input:} Similarity matrix $\mS$, current clustering $\vc^i$, concentration parameter $\beta$.
\STATE $\emM_{uk} \coloneqq -\sum_{v: c^i_v = k} S_{uv},\ \forall u \in \mathcal{V}, k \in [K]$
\STATE $\mQ,\mM \coloneqq$ \hyperref[alg:mf]{MeanField}$(\mS,\mM,\beta)$
\STATE Initialize $a^{\text{JEIG}}(u,v) \coloneqq 0,\ \forall (u,v) \in \mathcal{E}$
\FOR{$i \coloneqq 1$ to $m$}
    \STATE $\mathcal{D}_i \coloneqq \text{SelectPairs}(\mathcal{E})$ \COMMENT{$\mathcal{D}_i \subseteq \mathcal{E}$}
    \FOR{$j \coloneqq 1$ to $n$} 
        \STATE Sample $\ve \sim P(\rmE_{\mathcal{D}_i})$
        \STATE $\mQ^{(\mS_{\mathcal{D}_i} = \ve)} \coloneqq$ \hyperref[alg:mf]{MeanField}$(\mS,\mM,\beta \mid \mS_{\mathcal{D}_i} = \ve)$
        \STATE $a^{\text{JEIG}}(u,v) \mathrel{+}= \frac{1}{n}H(\ermE_{uv}\mid \mQ^{(\mS_{\mathcal{D}_i} = \ve)}),\ \forall (u,v) \in \mathcal{E}$
    \ENDFOR
\ENDFOR
\STATE $a^{\text{JEIG}}(u,v) \coloneqq H(\ermE_{uv}\mid\mQ) - \frac{1}{m}a^{\text{JEIG}}(u,v),\ \forall (u,v) \in \mathcal{E}$
\STATE \textbf{Return} $a^{\text{JEIG}}$
\end{algorithmic}
\end{algorithm}

%
\subsubsection{Joint Expected Information Gain} \label{section:JEIG}

In this section, we study the information gains defined in Eqs.~\ref{eq:igo2} and \ref{eq:JEIG2}. We approximate the conditional entropy

\begin{equation}
H(\ermE_{uv} \mid \rmE) = \mathbb{E}_{\ve \sim P(\rmE)}[H(\ermE_{uv} \mid \rmE = \ve)]
\end{equation}

in Eq.~\ref{eq:JEIG2} using the conditional mean-field approximation $\mQ^{(\mS_{\mathcal{E}} = \ve)}$. The conditional entropy

\begin{equation}
H(\ermE_{uv} \mid \rvy) = \mathbb{E}_{\vc \sim Q(\rvy)}[H(\ermE_{uv} \mid \rvy = \vc)]
\end{equation}

in Eq.~\ref{eq:igo2} is less straightforward. A natural approach is to update $\mS$ based on $\vc$ by setting $\emS_{uv} = +1$ if $\evc_u = \evc_v$ and $\emS_{uv} = -1$ otherwise, and then computing the corresponding conditional mean-field approximation. In both cases, we then approximate the entropy of $\ermE_{uv}$ given a mean-field approximation conditioned on all (or a subset) of the similarities being updated. Given this, we now derive a general estimator based on the information gain:
\begin{equation} \label{eq:igsub}
I(\ermE_{uv};\rmE_{\mathcal{D}}) = H(\ermE_{uv}) - H(\ermE_{uv} \mid \rmE_{\mathcal{D}}),
\end{equation}
where $\rmE_{\mathcal{D}} = \{\ermE_{uv} \mid (u,v) \in \mathcal{D}\}$ for some $\mathcal{D} \subseteq \mathcal{E}$. From the discussion above, the expressions in Eq. \ref{eq:igo2} and Eq. \ref{eq:JEIG2} can be seen as special cases of Eq. \ref{eq:igsub}. The entropy $H(\ermE_{uv})$ is approximated as in Eq.~\ref{eq:entropy}, while the conditional entropy is given by
\begin{equation} \label{eq:igig}
\begin{aligned}
&H(\ermE_{uv} \mid \rmE_{\mathcal{D}}) = \mathbb{E}_{\ve \sim P(\rmE_{\mathcal{D}})}[H(\ermE_{uv} \mid \rmE_{\mathcal{D}} = \ve)] \\
&= \sum_{\ve \in \{-1,+1\}^{|\mathcal{D}|}} P(\rmE_{\mathcal{D}} = \ve) H(\ermE_{uv} \mid \rmE_{\mathcal{D}} = \ve).
\end{aligned}
\end{equation}
We approximate $H(\ermE_{uv} \mid \rmE_{\mathcal{D}} = \ve)$ by

\begin{equation}
H(\ermE_{uv} \mid \mQ^{(\mS_{\mathcal{D}} = \ve)}) \triangleq -\sum_{e \in \{-1,+1\}} p(e) \log p(e),
\end{equation}

where $p(e) \triangleq P(\ermE_{uv} = e \mid \mQ^{(\mS_{\mathcal{D}} = \ve)})$. In this manner, we capture the joint impact of the pairs in $\mathcal{D}$ on the entropy of $\ermE_{uv}$.

Since the sum in Eq.~\ref{eq:igig} becomes intractable for large $|\mathcal{D}|$, we estimate it via Monte Carlo sampling. For generality, assume we have $m$ subsets $\mathcal{D}_1,\ldots,\mathcal{D}_m \subseteq \mathcal{E}$ and, for each subset, $n$ samples $\ve_i^1,\ldots,\ve_i^n \sim P(\rmE_{\mathcal{D}_i})$. Then, we define the acquisition function
\begin{equation} \label{eq:jeig}
a^{\text{JEIG}}(u,v) \triangleq H(\ermE_{uv}) - \frac{1}{mn} \sum_{i=1}^m \sum_{l=1}^n H(\ermE_{uv} \mid \mQ^{(\mS_{\mathcal{D}_i} = \ve_i^l)}).
\end{equation}
This formulation requires only $mn$ executions of Alg. \ref{alg:mf} and, in practice, small values of $m$ and $n$ yield good performance. Moreover, using multiple small subsets (with $|\mathcal{D}_i| \ll |\mathcal{E}|$) offers several benefits: (i) the Monte Carlo estimation is more accurate for smaller subsets, reducing the number of mean-field computations; (ii) If $\mathcal{D}_i = \mathcal{E}$, the conditional mean-field approximation $\mQ^{(\mS_{\mathcal{D}_i} = \ve)}$ is computed based on a similarity matrix where all pairs $(u, v) \in \mathcal{E}$ are sampled from $\emS_{uv} \sim P(\ermE_{uv})$, which will lead to extreme selection bias for the following reason: The probability $P(\ermE_{uv})$ (which is computed using $\mQ$) may already be biased (in particular in early iterations when $\mS$ contains incomplete/wrong information). Then, running Alg. \ref{alg:mf} from scratch with a new similarity matrix fully augmented with biased information, will exaggerate the bias further; (iii) Using $m$ different subsets makes the estimator in Eq. \ref{eq:jeig} generic and flexible, but also captures more information about each $\ermE_{uv}$, while remaining efficient and avoiding exaggerated selection bias.


Alg. \ref{alg:JEIG} outlines how $a^{\text{JEIG}}$ is computed. The algorithm first initializes $\mQ$ and $\mM$ via Alg. \ref{alg:mf}, then iterates $m$ times. In each iteration, a subset $\mathcal{D}_i \subseteq \mathcal{E}$ is selected, and a Monte Carlo estimation of Eq. \ref{eq:igig} is performed (lines 7--10). Various strategies for selecting $\mathcal{D}_i$ yield good performance; see Section~\ref{section:exp-setup} for details.


\section{Experiments} \label{section:experiments}

In this section, we describe our experimental studies. Each active learning procedure was executed on 1 core of an Intel(R) Xeon(R) Gold 6338 CPU @ 2GHz (with 32 cores total). We have access to a compute cluster with many of these CPU's allowing us to execute many procedures in parallell. Each CPU has access to 128GB of RAM (shared among cores), but much less would suffice for our experiments. 

\subsection{Experimental Setup} \label{section:exp-setup}



\noindent\textbf{Datasets.} Below is a description of the eight datasets used in our experiments. Datasets 2–6 are obtained from the UCI Machine Learning Repository \cite{uci}. For all datasets, a random subset of at most $N = 1000$ objects are considered for the active CC experiments (due to high runtime of EIG-O and some of the baseline methods).

\noindent\textbf{Correlation clustering algorithm.} We use the local search CC algorithm proposed by \cite{anonymous} (see Alg. \ref{alg:clustering-alg}) on line 4 of Alg. \ref{alg:acc}. It is highly robust to noise in $\mS$ and dynamically determines the number of clusters.

\noindent\textbf{Ground-truth similarities.} For each experiment, we are given a dataset $\mathbf{X}$ with ground-truth labels $\vc^{\ast}$, where the ground-truth labels are only used for evaluations. Then, for each $(u, v) \in \mathcal{E}$ in a dataset, we set $S^{\ast}_{uv}$ to $+1$ if $u$ and $v$ belong to the same cluster, and $-1$ otherwise.

\noindent\textbf{Oracles.} We investigate two different oracles in Alg. \ref{alg:acc}: (i) \textbf{Oracle 1}. Returns $\emS_{uv}^{\ast}$ with probabiltiy $1-\gamma$ or a uniform random value in $[-1, +1]$ with probability $\gamma$; (ii) \textbf{Oracle 2}. We split the dataset into two disjoint parts $\mathbf{X} = \mathbf{X}_{\text{train}} \cup \mathbf{X}_{\text{test}}$. Then, we train a pairwise prediction model $f_{\theta} : \mathbf{X} \times \mathbf{X} \rightarrow [-1, +1]$ on $\mathbf{X}_{\text{train}}$, where the ground-truth similarities $\mS^{\ast}$ are used as labels. Given any two data points $\mathbf{x}_u, \mathbf{x}_v \in \mathbf{X}$, we can predict their similarity as $f_{\theta}(\mathbf{x}_u, \mathbf{x}_v) \in [-1, +1]$. We then perform the CC experiments on data points in $\mathbf{X}_{\text{test}}$, and the oracle always returns the similarity $f_{\theta}(\mathbf{x}_u, \mathbf{x}_v)$. The ground-truth similarities of data points in $\mathbf{X}_{\text{test}}$ are never used when training $f_{\theta}$. The motivation for these oracles are as follows. Oracles 1 corresponds to cases where the oracle provides unbiased information about $\mS^{\ast}$ (but noisy), allowing recovery of the ground-truth clustering solution $\vc^{\ast}$. This 
is considered by previous work \cite{mazumdar2017-2, silwal2023kwikbucks, anonymous}. Oracle 2 may provide biased similarities due to noise/ambiguity in the feature space, and exact recovery of $\vc^{\ast}$ may not be possible. This method is suggested by, e.g., \cite{BansalBC04,silwal2023kwikbucks} to compute pairwise similarities for CC. 

\noindent\textbf{Initial similarities.} Before running the active learning procedure in Alg. \ref{alg:acc}, we randomly sample an initial set of $B_0$ similarities. The value of $B_0$ for each dataset (chosen based on the dataset size $N$) is provided in the list of datasets below. Initial random sampling is a common strategy in active learning to avoid selection bias, and it does not limit our methods, as it can be always be done in practice.


\noindent\textbf{Repeated queries.} In general, Alg. \ref{alg:acc} supports multiple queries for the same pairwise similarity. This assumes each query for the same pair provides more information about the underlying distribution, which would be applicable to oracle 1. This is a common approach in active learning to deal with noisy oracles \cite{requery, settles.tr09}. However, \textit{we do not consider multiple queries for the same pair in our experiments}, as we found the difference in performance to be very small.

\noindent\textbf{Acquisition functions.} We have introduced three novel acquisition functions in this paper: $a^{\text{Entropy}}$ (Eq. \ref{eq:entropy}), $a^{\text{EIG-O}}$ (Eq. \ref{eq:eigo}), and $a^{\text{JEIG}}$ (Eq. \ref{eq:jeig}). We compare these methods with \emph{maxexp} and \emph{maxmin} from \cite{anonymous}. In short, both \emph{maxmin} and \emph{maxexp} aim to query pairs with small absolute similarity that belong to triples $(u, v, w)$ that violate the transitive property of pairwise similarities. In other words, the goal is to reduce the inconsistency of $\mS$ by resolving violations of the transitive property in triples. We include a simple baseline $a^{\text{Uniform}}(u, v) \sim \text{Uniform}(0, 1)$ which selects pairs uniformly at random. \cite{anonymous} compares maxexp/maxmin to a pivot-based active CC algorithm called {QECC} \cite{bonchi2020} and two adapted state-of-the-art active constraint clustering methods, called {COBRAS} \cite{Craenendonck2018COBRAAF} and {nCOBRAS} \cite{lirias3060956}. We also include them here. Finally, we include a recent active CC approach based on multi-armed bandits called KC-FB \cite{noisyqecc}. Unlike the active CC procedure described in Alg. \ref{alg:acc}, QECC, COBRAS, nCOBRAS, and KC-FB cannot be updated incrementally with a fixed batch size. For a fair comparison, we therefore restart each of these algorithms from scratch after every query round, assigning them a fixed budget equal to the total number of queries made up to that point.


\noindent\textbf{Batch diversity.} In this paper, we consider single-sample acquisition functions that do not explicitly consider the joint informativeness among the elements in a batch $\mathcal{B}$. This has the benefit of avoiding the combinatorial complexity of selecting an optimal batch, which is a common problem for batch active learning \cite{deep-al-survey}. However, the work in \cite{kirsch2023stochastic} proposes a simple method for improving the batch diversity for single-sample acquisition functions using noise. In this paper, we utilize the \textit{power} acquisition method. Given some acquisition function $a^{\text{X}}$, this corresponds to $a^{\text{PowerX}}(u, v) = \log (a^{\text{X}}(u, v)) + \epsilon_{uv}$ where $\epsilon_{uv} \sim \text{Gumbel}(0;1)$. This is used for all information-theoretic acquisition functions proposed in this paper. We observe no benefit of this for maxmin/maxexp, likely due to their inherent randomness. Finally, it is not applicable to the other baselines, as they do not follow Alg. \ref{alg:acc}, as described earlier.

\noindent\textbf{Hyperparameters.} Unless otherwise specified, the following hyperparameters are used. The batch size $B$ depends on the dataset (since each dataset is of different size $N$). The value of $B$ for each dataset is provided in the list of datasets below. For all information-theoretic acquisition functions (which depend on $P^{\text{Gibbs}}$, see Eq. \ref{eq:gibbs}), we set $\beta = 3$. For $a^{\text{EIG-O}}$, we set $|\mathcal{E}^{\text{EIG}}| = 20N$. For $a^{\text{JEIG}}$, we select the top-$|\mathcal{D}_i|$ pairs according to $\log(a^{\text{Entropy}}(u, v)) + \epsilon_{uv}$ where $\epsilon_{uv} \sim \text{Gumbel}(0;1)$. In other words, the top-$|\mathcal{D}_i|$ pairs according to $a^{\text{Entropy}}$ with some added acquisition noise (as explained above). This leads to diversity among the selected $\mathcal{D}_i$, while containing pairs with large entropy. Pairs with large entropy are likely to have large impact on each $\ermE_{uv}$, and are therefore important to include in each $\mathcal{D}_i$. We set $|\mathcal{D}_i| = 0.02|\mathcal{E}|$ (i.e., 2\% of all pairs), $m = 5$ and $n = 50$ for all datasets. Finally, see Section \ref{section:sensitivity} for a detailed analysis and discussion of different hyperparameters.

\noindent\textbf{Performance evaluation.} In each iteration of the active CC procedure (Alg. \ref{alg:acc}), we calculate the Adjusted Rand Index (ARI) between the current clustering $\vc^i$ and the ground truth clustering $\vc^{\ast}$ (i.e., ground truth labels of dataset). Intuitively, ARI measures how similar the two clusterings are, where a value of 1 indicates they are identical. Each active learning procedure is repeated 10 times with different random seeds, where the standard deviation is indicated by a shaded color or an error bar.

\noindent\textbf{Datasets.} Below is a description of the eight datasets used in our experiments. Datasets 2–6 are obtained from the UCI Machine Learning Repository \cite{uci}. For all datasets, a random subset of at most $N = 1000$ objects are considered for the active CC experiments (due to high runtime of EIG-O and some of the baseline methods).

\begin{enumerate}[leftmargin=18pt, label=(\arabic*)]
    \item \textbf{CIFAR10} \cite{cifar10}: We use a random subset of $N=1000$ images (with $|\mathcal{E}| = 499500$) with cluster sizes: [91, 96, 107, 89, 99, 113, 96, 93, 112, 104]. A ResNet18 model \cite{resnet} trained on the full CIFAR10 dataset is used to embed images into a 512-dimensional space. For Oracle 2, $f_{\theta}$ is trained on these latent representations. We set the initial edge subset size $B_0 = 2500$ and the batch size $B = 1250$.  

    \item \textbf{20newsgroups}: This dataset consists of 18846 newsgroup posts across 20 topics. We focus on 5 topics: "rec.sport.baseball", "soc.religion.christian", "rec.autos", "talk.politics.mideast", and "misc.forsale". A random sample of $N=1000$ posts is used (with $|\mathcal{E}| = 499500$) with cluster sizes: [201, 190, 201, 217, 191]. Each document is embedded into a 768-dimensional space using the "distilbert-base-uncased" transformer model from the Flair Python library \cite{akbik2018coling}. For Oracle 2, $f_{\theta}$ is trained on the latent representations. We set $B_0 = 2500$ and $B = 250$.
    
    \item \textbf{Cardiotocography}: This dataset includes 2126 fetal cardiotocograms with 22 features and 10 classes. We use a sample of $N=1000$ data points (with $|\mathcal{E}| = 499500$) with cluster sizes: [180, 275, 27, 35, 31, 148, 114, 62, 28, 100]. We set $B_0 = 2500$ and $B = 750$.
    
    \item \textbf{Ecoli}: A biological dataset on protein cellular localization, containing $N=336$ samples across 8 clusters (with $|\mathcal{E}| = 56280$). The cluster sizes are: [137, 76, 1, 2, 37, 26, 5, 52]. We set $B_0 = 280$ and $B = 85$.
    
    \item \textbf{Forest Type Mapping}: A remote sensing dataset of 523 samples collected from Japanese forests, grouped into 4 forest types (with $|\mathcal{E}| = 136503$). Cluster sizes are: [168, 84, 86, 185]. We set $B_0 = 500$ and $B = 350$.
    
    \item \textbf{User Knowledge Modelling}: This dataset contains the knowledge status of 403 students on Electrical DC Machines, grouped into 5 classes (with $|\mathcal{E}| = 81003$). The cluster sizes are: [111, 129, 116, 28, 19]. We set $B_0 = 400$ and $B = 200$.

    \item \textbf{MNIST} \cite{DBLP:journals/pieee/LeCunBBH98}: We use a sample of $N=1000$ images (with $|\mathcal{E}| = 499500$) with cluster sizes: [105, 109, 111, 112, 104, 86, 99, 88, 88, 98]. A simple CNN trained on MNIST is used to embed the images into a 128-dimensional space. For Oracle 2, $f_{\theta}$ is trained on these embeddings. We set $B_0 = 2500$ and $B = 1250$.  
    
    \item \textbf{Synthetic}: The synthetic dataset consists of 500 normally distributed 10-dimensional data points evenly split into 10 clusters (with $|\mathcal{E}| = 124750$). We set $B_0 = 500$ and $B = 300$.
\end{enumerate}


\subsection{Results}

\begin{figure*}[t]
  \centering
  \includegraphics[width=\textwidth]{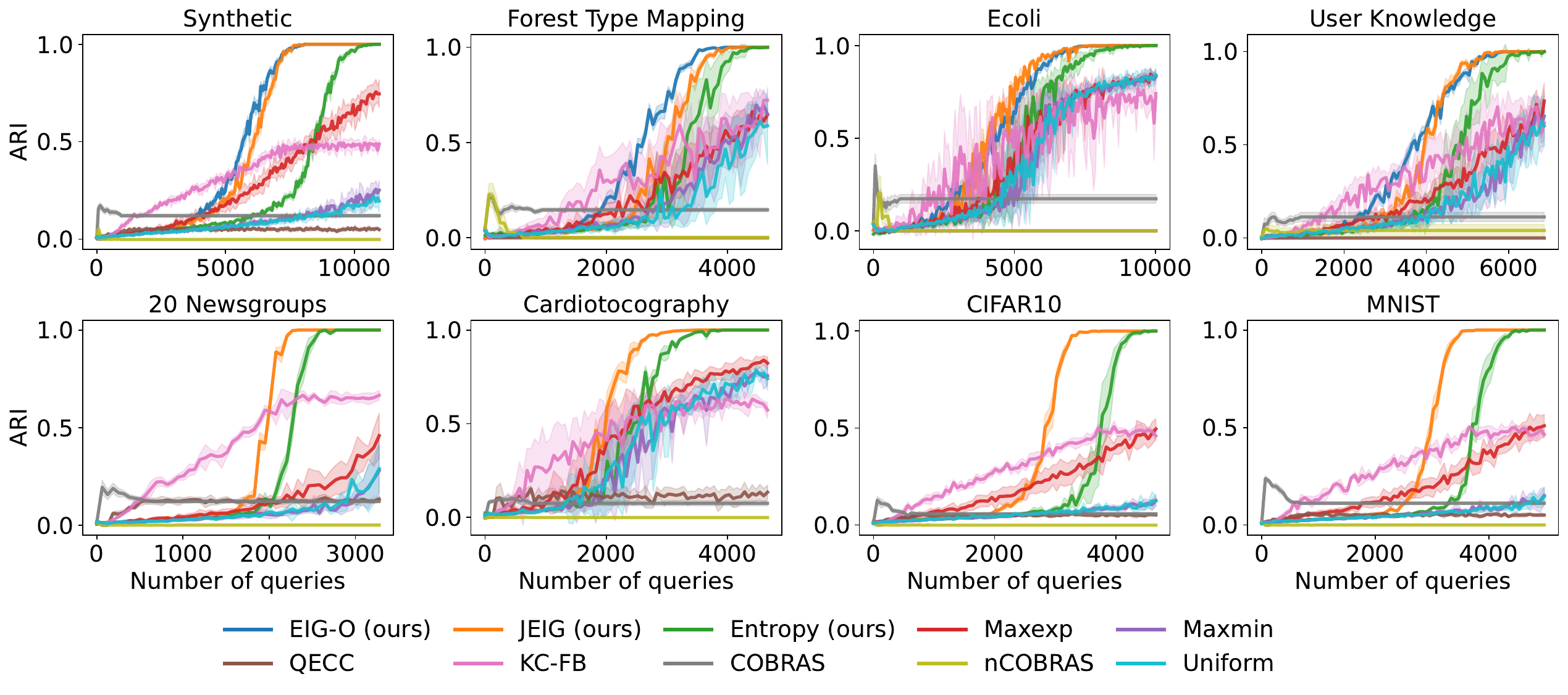}
  \vspace{-1em}
  \caption{Results for \textbf{Oracle 1} with noise level $\gamma = 0.4$. The evaluation metric is the adjusted rand index (ARI).}
  \label{fig:ari-ground-truth}
\end{figure*}

\begin{figure*}[t]
  \centering
  \includegraphics[width=\textwidth]{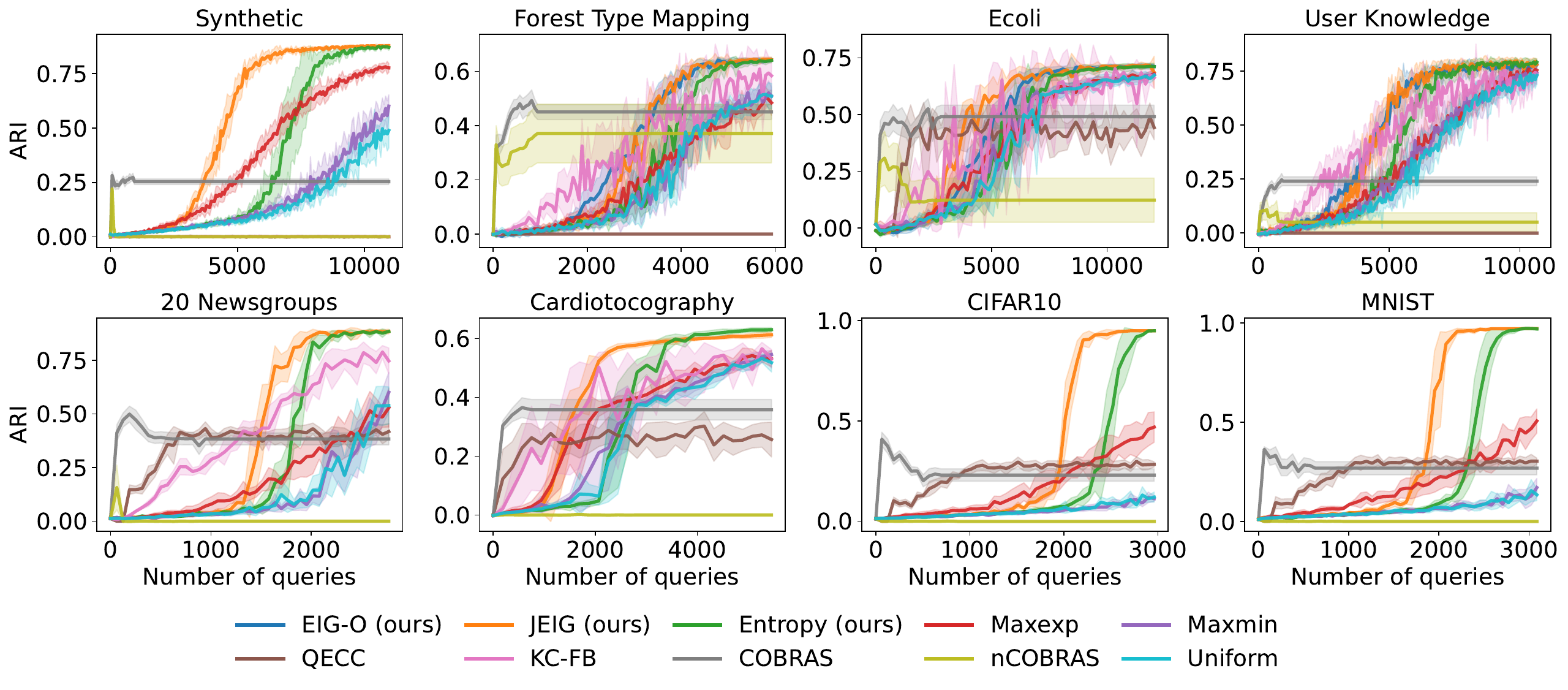}
  \vspace{-1em}
  \caption{Results for \textbf{Oracle 2}. The evaluation metric is the adjusted rand index (ARI).}
  \label{fig:ari-predict-gt}
\end{figure*}

\begin{figure*}[t!]
\centering 
\subfigure[Varying noise level\label{fig:noiselevel}]
{\includegraphics[width=0.24\linewidth]{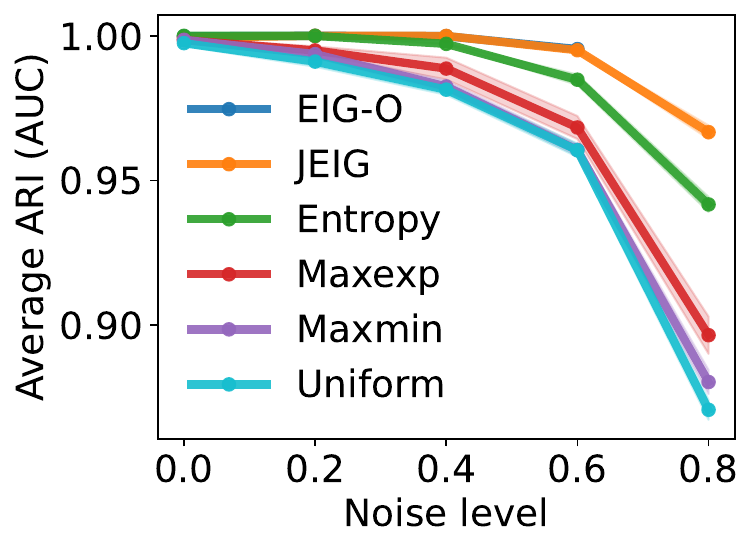}} 
\subfigure[Varying $|\mathcal{E}^{\text{EIG}}|$ for $a^{\text{EIG-O}}$\label{fig:eigo}]
{\includegraphics[width=0.24\linewidth]{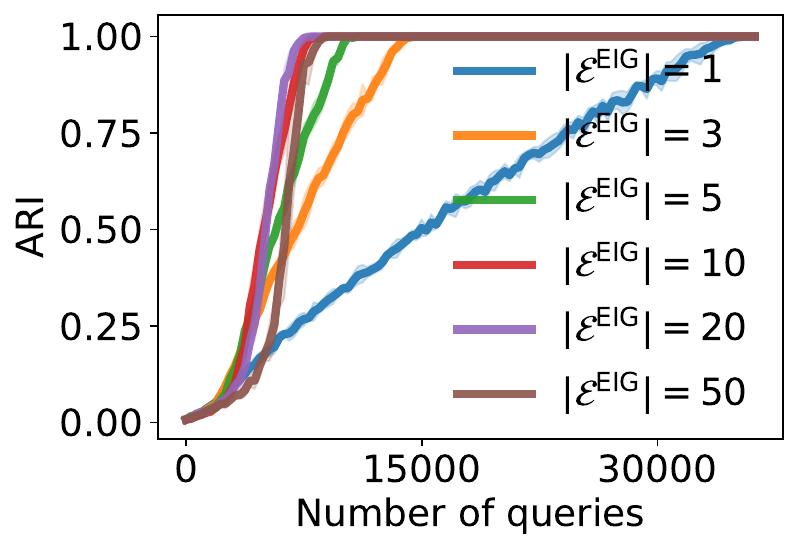}}
\subfigure[Varying $|\mathcal{D}_i|$ for $a^{\text{JEIG}}$\label{fig:jeig}]
{\includegraphics[width=0.24\linewidth]{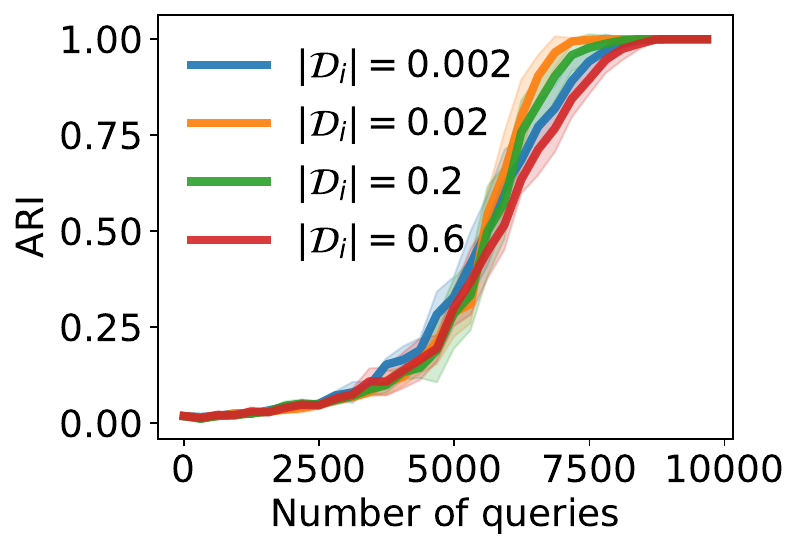}}
\subfigure[Varying $\beta$ for $a^{\text{Entropy}}$\label{fig:entropy}]
{\includegraphics[width=0.24\linewidth]{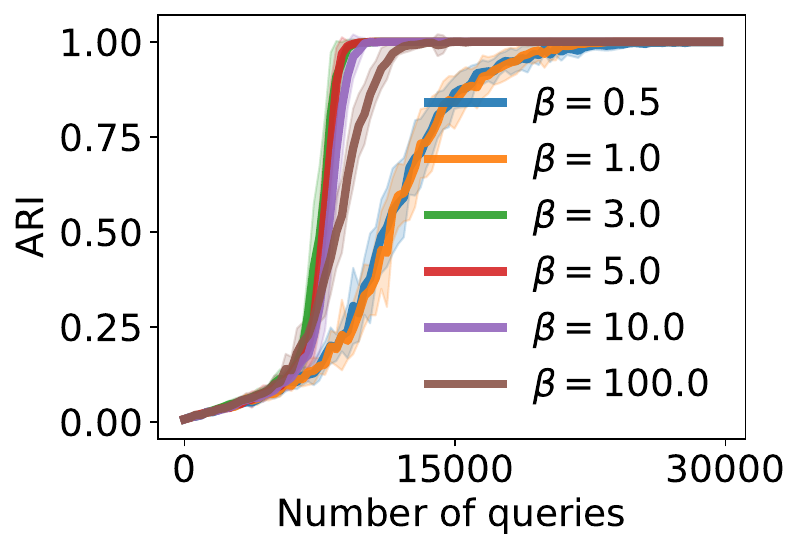}}
\caption{(a) Varying the noise level $\gamma$ of various methods on the synthetic dataset using Oracle 1. Some baselines are excluded for clarity due to very poor performance. (b) Varying the number of candidate pairs $|\mathcal{E}^{\text{EIG}}|$ for $a^{\text{EIG-O}}$. (c) Varying the subset size $|\mathcal{D}_i|$ for $a^{\text{JEIG}}$. (d) Varying the concentration parameter $\beta$ for $a^{\text{Entropy}}$.}
\label{fig:sensitivity}
\end{figure*}

Figures \ref{fig:ari-ground-truth}-\ref{fig:ari-predict-gt} show the results for different datasets for Oracle 1 and Oracle 2, respectively. We observe that all the information-theoretic acquisition functions introduced in this paper significantly outperform the baseline methods. In addition, the acquisition functions based on information gain ($a^{\text{EIG-O}}$ and $a^{\text{JEIG}}$) consistently outperform $a^{\text{Entropy}}$. This indicates the effectiveness of augmenting the similarity matrix $\mS$ with pseudo-similarities predicted by the current model $\mQ$ as $\emS_{uv} \sim P(\ermE_{uv} \mid \mQ)$, before quantifying the model uncertainty. The acquisition functions based on information gain perform rather similarly. This is due to the fact that they are based on closely connected quantities (as described in Section \ref{section:ig}). However, $a^{\text{JEIG}}$ is consistently among the best performing acquisition functions, while also being more computationally efficient compared to $a^{\text{EIG-O}}$ (see Section \ref{section:runtime} for an investigation of the runtimes of all methods). Because of this, we exclude $a^{\text{EIG-O}}$ in some cases due to its computational inefficiency. 

Both \emph{maxmin} and \emph{maxexp} perform significantly worse. This is likely because they spend too many queries with the goal of resolving the inconsistency of $\mS$. However, the CC algorithm used (Alg. \ref{alg:clustering-alg}) is robust to inconsistency in $\mS$. The remaining baselines perform poorly under Oracle 1, suggesting a high sensitivity to noise. Notably, COBRAS and nCOBRAS are the only methods that leverage the feature vectors. As shown in Figure \ref{fig:ari-predict-gt}, both methods perform well initially but eventually converge to suboptimal solutions. This behavior is likely due to noise or ambiguity in the feature space, which disproportionately affects methods that rely heavily on this information. In contrast, the other methods, including ours, are less impacted, as they do not overemphasize the feature space (they rely solely on information coming from the oracle in terms of pairwise similarities).

\subsection{Sensitivity Analysis} \label{section:sensitivity}

\begin{figure}[t]
  \centering
  \includegraphics[width=1\linewidth]{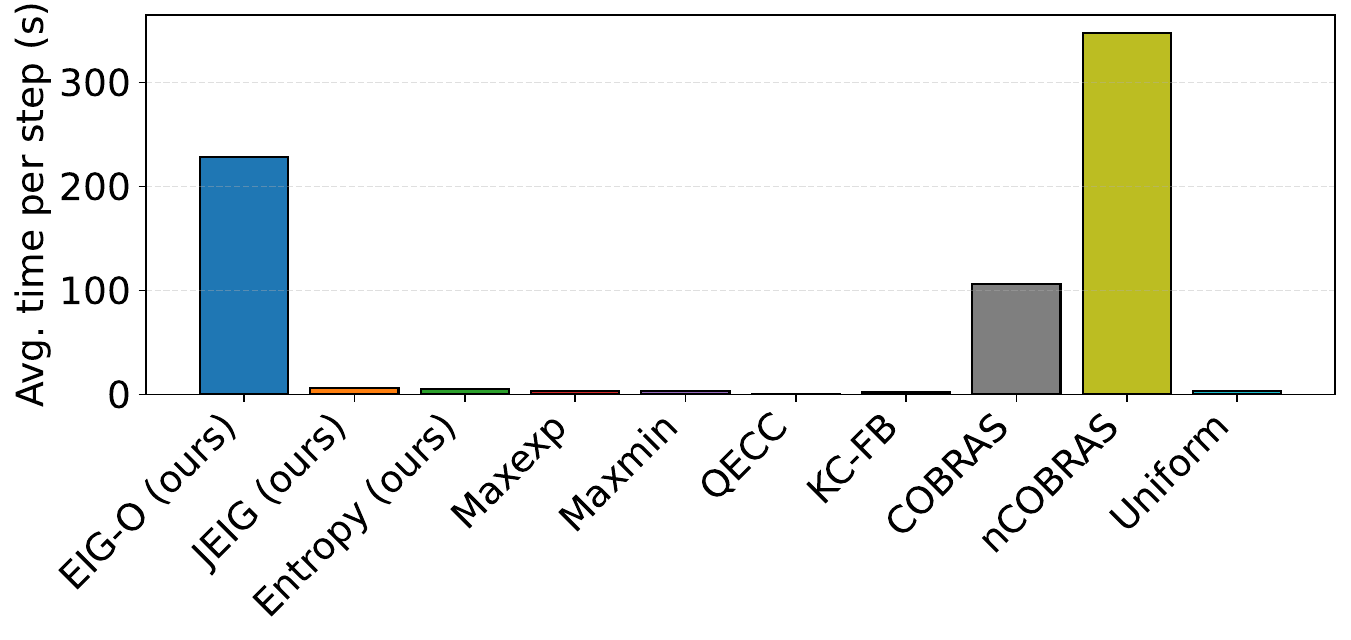}
  \vspace{-1em}
  \caption{Average runtime of each iteration across four of the datasets (synthetic, forest type mapping, ecoli and user knowledge) in seconds.}
  \label{fig:runtime}
\end{figure}


We here investigate the sensitivity of acquisition functions when varying various parameters. All the results in this section are performed on the synthetic dataset using Oracle 1.  Figures \ref{fig:noiselevel}-\ref{fig:entropy} respectively show the results when varying the noise level $\gamma$, $|\mathcal{E}^{\text{EIG}}|$ for $a^{\text{EIG-O}}$, $|\mathcal{D}_i|$ for $a^{\text{JEIG}}$ and $\beta$ for $a^{\text{Entropy}}$. For Figure \ref{fig:noiselevel}, the $y$-axis corresponds to the area under the curve (AUC) of the active learning plot w.r.t. the respective performance metric (i.e., ARI) where higher is better.

We see that our acquisition functions are very robust to noise. In addition, the benefit of our proposed acquisition functions increases with larger noise levels. This is consistent with previous work on active learning, where the benefit of many acquisition functions over uniform selection increases as the complexity of the problem increases. As expected, increasing the size of \( |\mathcal{E}^{\text{EIG}}| \) improves performance, as it allows \( a^{\text{EIG-O}} \) to be computed for a larger number of pairs. In contrast, the performance is sensitive to the choice of \( |\mathcal{D}_i| \); both overly small and excessively large values can worsen performance. This aligns with the discussion in Section~\ref{section:JEIG}: a large \( |\mathcal{D}_i| \) may introduce significant selection bias, while a small one may fail to capture sufficient information. Lastly, we observe that setting \( \beta = 3 \) yields strong performance, a trend that holds consistently for both \( a^{\text{EIG-O}} \) and \( a^{\text{JEIG}} \).

\subsection{Runtime} \label{section:runtime}


In Figure~\ref{fig:runtime}, we present the average runtime per iteration (in seconds) for all acquisition functions, averaged over the four datasets for which results were available for all methods. We observe that \( a^{\text{Entropy}} \) is highly efficient, with runtime comparable to other baseline methods. Among the information gain–based acquisition functions, \( a^{\text{JEIG}} \) is the most efficient and performs close to \( a^{\text{Entropy}} \). As expected, \( a^{\text{EIG-O}} \) is the least efficient. Overall, both \( a^{\text{JEIG}} \) and \( a^{\text{Entropy}} \) offer competitive runtime performance while significantly outperforming the baselines in clustering quality relative to the number of queries.

\section{Conclusion}

We introduced a family of information-theoretic acquisition functions for active correlation clustering, including \textit{Entropy}, \textit{EIG-O}, and \textit{JEIG}, based on principled uncertainty estimation via mean-field approximation. Our methods significantly outperform existing baselines across multiple datasets and oracle types, with \textit{JEIG} offering the best balance of effectiveness and efficiency. These results demonstrate the benefits of leveraging model uncertainty in active clustering, particularly in noisy settings, and open up new directions for principled query selection in non-parametric clustering frameworks.

\section*{Acknowledgments}

This work was partially supported by the Wallenberg AI, Autonomous Systems and Software Program (WASP) funded by the Knut and Alice Wallenberg Foundation. The computations and data handling was enabled by resources provided by the National Academic Infrastructure for Supercomputing in Sweden (NAISS), partially funded by the Swedish Research Council through grant agreement no. 2022-06725.

\bibliographystyle{IEEEtran}
\bibliography{IEEEabrv, references}

\appendices
\section{Proofs} \label{appendix:proofs}
\propmaxcorr*
\begin{proof}
As described in \cite{ethz-a-010077098,Chehreghani22_shift}, we can write the cost function in Eq. \ref{eq:cost} as

\begin{align} \label{eq:CC2MaxCor}
&R^{\text{CC}}(\vc \mid \mS) =\sum_{(u, v) \in \mathcal{E}} V(u, v \mid \mS, \vc) \nonumber \\
&\quad= \sum_{\substack{(u,v) \in \mathcal{E}\\ c_u = c_v}}\frac{1}{2} (|S_{uv}|-S_{uv}) \nonumber + \sum_{\substack{(u, v) \in \mathcal{E}\\ c_u \ne c_v}}\frac{1}{2} (|S_{uv}|+S_{uv}) \nonumber \\
&\quad= \frac{1}{2} \sum_{(u,v) \in \mathcal{E}} |S_{uv}| - \frac{1}{2}\sum_{\substack{(u,v) \in \mathcal{E}\\ c_u = c_v}} S_{uv} \nonumber \\
&+ \frac{1}{2} \sum_{(u,v) \in \mathcal{E}} S_{uv} - \frac{1}{2}\sum_{\substack{(u,v) \in \mathcal{E}\\ c_u = c_v}} S_{uv} \nonumber \\
&\quad= \underbrace{\frac{1}{2} \sum_{(u,v) \in \mathcal{E}} (|S_{uv}| + S_{uv})}_{\text{constant}} - \sum_{\substack{(u,v) \in \mathcal{E}\\ c_u = c_v}} S_{uv}.
\end{align}

The first term in Eq. \ref{eq:CC2MaxCor} is \emph{constant} w.r.t. the choice of a particular clustering $\vc$. 

\end{proof}

\propkl*

\begin{proof}

Given our cost function $R^{\text{MC}}$ (Eq. \ref{eq:maxcorrfn}), the \emph{generalized free energy} is defined as \cite{DBLP:journals/pami/HofmannPB98}

\begin{equation}
\begin{aligned}
    \mathcal{F}_{\beta}(P) &\triangleq \mathbb{E}_{P(\rvy)}[R^{\text{MC}}(\rvy)] - \frac{1}{\beta}H(P) \\
    &= \sum_{\vc \in \mathcal{C}} P(\vc) R^{\text{MC}}(\vc) + \frac{1}{\beta}\sum_{\vc \in \mathcal{C}} P(\vc) \log P(\vc),
\end{aligned}
\end{equation}

for some $P \in \mathcal{P}$ where $\mathcal{P}$ is the set of distributions with sample space $\mathcal{C}$. The Gibbs distribution $P^{\text{Gibbs}}$ minimizes the generalized free energy \cite{DBLP:journals/pami/HofmannPB98} and is called the \emph{free energy}. It can be written as

\begin{equation}
    \mathcal{F}_{\beta}(P^{\text{Gibbs}}) = -\frac{1}{\beta} \log \mathcal{Z},
\end{equation}

where $\mathcal{Z} \triangleq \sum_{\vc^{\prime} \in \mathcal{C}} \exp(-\beta R^{\text{MC}}(\vc^{\prime}))$ is the normalizing constant of the Gibbs distribution in Eq. \ref{eq:gibbs}. Given this, we can now simplify the KL-divergence.

\begin{equation}
\begin{aligned}
&  \KL(Q\| P^{\text{Gibbs}}) =\sum_{\vc \in \mathcal{C}} Q(\vc) \log \frac{Q(\vc)}{P^{\text{Gibbs}}(\vc)} \\
& \quad =\sum_{c \in \mathcal{C}} Q(\vc) \log \frac{Q(\vc)}{\exp \left(-\beta\left( R^{\text{MC}}(\vc)-\mathcal{F}_{\beta}(P^{\text{Gibbs}})\right)\right)} \\
& \quad =\sum_{c \in \mathcal{C}} Q(\vc)\left[\log Q(\vc)+\beta\left( R^{\text{MC}}(\vc)-\mathcal{F}_{\beta}(P^{\text{Gibbs}})\right)\right] \\
&\quad =\sum_{u \in \mathcal{V}} \sum_{k \in [K]} Q_{uk} \log Q_{uk}+\beta \mathbb{E}_{Q(\rvy)}[ R^{\text{MC}}(\rvy)]-\beta \mathcal{F}_{\beta}(P^{\text{Gibbs}}) \\
&\quad = \beta \mathbb{E}_{Q(\rvy)}[ R^{\text{MC}}(\rvy)] - \sum_{u \in \mathcal{V}} H(\ervy_u) - \beta \mathcal{F}_{\beta}(P^{\text{Gibbs}}) \\
&\quad = \beta \mathcal{F}_{\beta}(Q) - \beta \mathcal{F}_{\beta}(P^{\text{Gibbs}}) \\
&\quad \geq 0,
\end{aligned}
\end{equation}

where $H(\ervy_u) \triangleq -\sum_{k \in [K]} \emQ_{uk} \log \emQ_{uk}$ is the entropy of $\ervy_u$. The last inequality is a property of the KL-divergence. From this, we have the bound

\begin{equation} \label{eq:bound}
\mathcal{F}_{\beta}(P^{\text{Gibbs}}) \leq \mathcal{F}_{\beta}(Q),
\end{equation}

and minimizing the KL-divergence corresponds to minimizing the generalized free energy $\mathcal{F}_{\beta}$ w.r.t. factorial distributions $Q \in \mathcal{Q}$, which is consistent with the maximum entropy principle. From this, minimizing the KL-divergence corresponds to the following optimization problem. 


\begin{align}
    & Q^* = \operatorname*{arg\,min}_{Q \in \mathcal{Q}} \mathcal{F}_{\beta}(Q)\nonumber \\
    & \text{s.t.} \quad \sum_{k \in [K]} Q_{uk} = 1 \quad \forall u \in \mathcal{V}. \label{eq:kl-constr}
\end{align}

Then, by applying a Lagrangian relaxation to the constraint in Eq. \ref{eq:kl-constr} and setting the gradient of the objective w.r.t. $Q_{uk}$ to zero, we obtain

\begin{equation}\label{eq:mf-first}
\begin{aligned}
0 & =\frac{\partial}{\partial Q_{uk}}\mathbb{E}_{Q(\rvy)}[R^{\text{MC}}(\rvy)] \\
&-\frac{1}{\beta}\sum_{v \in \mathcal{V}} H(\ervy_v)+\sum_{w \in \mathcal{V}} \mu_w(\sum_{k \in [K]} Q_{wk}-1) \\
& =\frac{\partial}{\partial Q_{uk}}\sum_{\vc \in \mathcal{C}} \prod_{v \in \mathcal{V}} Q_{vc_v}  R^{\text{MC}}(\vc) \\
&- \frac{1}{\beta}\sum_{v \in \mathcal{V}} H(\ervy_v)+\sum_{w \in \mathcal{V}} \mu_w(\sum_{k \in [K]} Q_{wk}-1) \\
& =\sum_{\vc \in \mathcal{C}} \prod_{\substack{v \in \mathcal{V} \\ v \neq u}} Q_{vc_v} \1_{\{c_u=k\}}  R^{\text{MC}}(\vc)+\frac{1}{\beta}\left(\log Q_{uk}+1\right)+\mu_u \\
&= \mathbb{E}_{Q(\rvy \mid \ervy_u = k)}[ R^{\text{MC}}(\rvy)]+\frac{1}{\beta}\left(\log Q_{uk}+1\right)+\mu_u,
\end{aligned}
\end{equation}

where $\mu_u$'s are the Lagrange multipliers and we define $\emM_{uk} \triangleq \mathbb{E}_{Q(\rvy \mid \ervy_u = k)}[ R^{\text{MC}}(\rvy)]$ as the mean-fields, which correspond to the expected cost subject to the constraint that object $u$ is assigned to cluster $k$. We can simplify 

\begin{equation}\label{eq:mf-simplification2}
\begin{aligned}
\emM_{uk} &=\mathbb{E}_{Q(\rvy \mid \ervy_u = k)}[ R^{\text{MC}}(\rvy)] \\ 
&= \mathbb{E}_{Q(\rvy \mid \ervy_u = k)}\left[-\sum_{(v, w) \in \mathcal{E}} S_{vw}\right] \\
&= \mathbb{E}_{Q(\rvy \mid \ervy_u = k)}\left[-\sum_{l \in [K]}\sum_{(v, w) \in \mathcal{E}} \1_{\{\ervy_v = l\}} \1_{\{\ervy_w = l\}} S_{vw}\right] \\
&= -\sum_{l \in [K]}\sum_{(v, w) \in \mathcal{E}} \mathbb{E}_{Q(\rvy \mid \ervy_u = k)}[\1_{\{\ervy_v = l\}} \1_{\{\ervy_w = l\}}] S_{vw} \\
&= -\sum_{l \in [K]}\sum_{(v, w) \in \mathcal{E}} S_{vw} Q_{vl}Q_{wl} \\
&= -\sum_{l \in [K]}\sum_{\substack{v \in \mathcal{V} \\ v \neq u}} S_{uv} Q_{ul}Q_{vl}-\sum_{l \in [K]}\sum_{\substack{(v, w) \in \mathcal{E} \\ v \neq u \\ w \neq u}} S_{vw} Q_{vl}Q_{wl} \\
&= -\sum_{\substack{v \in \mathcal{V} \\ v \neq u}} S_{uv} Q_{vk}-\underbrace{\sum_{l \in [K]}\sum_{\substack{(v, w) \in \mathcal{E} \\ v \neq u \\ w \neq u}} S_{vw} Q_{vl}Q_{wl}}_{\text{constant}}
\end{aligned}
\end{equation}

where the last equality uses that $Q_{ul} = 1$ if $l = k$ and $0$ otherwise, according to $Q(\vc \mid c_u = k)$. The second term of the last expression is a constant w.r.t. $Q_{uk}$ and is thus irrelevant for optimization (since it does not depend on $u$).

With the definition of $\emM_{uk}$, we can rewrite Eq. \ref{eq:mf-first} as

\begin{equation}\label{eq:mf-second}
\begin{aligned}
0 & = \emM_{uk}+\frac{1}{\beta}\left(\log Q_{uk}+1\right)+\mu_u.
\end{aligned}
\end{equation}

Then, we have
\begin{equation}\label{eq:mf-third}
\begin{aligned}
& \log Q_{uk} = -\beta \emM_{uk}- \beta \mu_u \\
\Rightarrow	 & Q_{uk} = \exp{(-\beta \emM_{uk})}\exp{(-\beta \mu_u)}.
\end{aligned}
\end{equation}

On the other hand, we have: $\sum_{k'} Q_{uk'} = 1$. Therefore,

\begin{equation}\label{eq:mf-forth}
\begin{aligned}
&\sum_{k'} \log Q_{uk'} = \sum_{k'}{\exp{(-\beta \emM_{uk'})}\exp{(-\beta \mu_u)}} =1 \\
\Rightarrow	 & \exp{(-\beta \mu_u)} = \frac{1}{\sum_{k'}{\exp{(-\beta \emM_{uk'})}}} \, .
\end{aligned}
\end{equation}

Then, inserting Eq. \ref{eq:mf-forth} into Eq. \ref{eq:mf-third} yields

\begin{equation}\label{eq:mf-fifth}
\begin{aligned}
Q_{uk} = \frac{\exp{(-\beta \emM_{uk})}}{\sum_{k'}{\exp{(-\beta \emM_{uk'})}}} \, .
\end{aligned}
\end{equation}

This derivation suggest an EM-type procedure for minimizing the KL-divergence $\KL(Q\| P^{\text{Gibbs}})$, which consists of alternating between estimating $Q_{uk}$'s given $\emM_{uk}$'s and then updating $\emM_{uk}$'s given the new values of $Q_{uk}$'s (as described in Alg. \ref{alg:mf}).

Finally, we can compute the Hessian of the objective as

\begin{equation}\label{eq:secondderiv}
\begin{aligned}
\frac{\partial^2}{\partial Q_{uk}^2} \mathcal{F}_{\beta}(Q) &= \frac{\partial}{\partial Q_{uk}} \emM_{uk}+\frac{1}{\beta}\left(\log Q_{uk}+1\right)+\mu_u \\
&= \frac{1}{\beta Q_{uk}} \\
& > 0.
\end{aligned}
\end{equation}

The positivity of the Hessian in Eq. \ref{eq:secondderiv} ensures that the generalized free energy $\mathcal{F}_{\beta}(Q)$ is convex with respect to $Q_{uk}$ for each object $u$, guaranteeing that the update for $Q_{uk}$ strictly decreases $\mathcal{F}_{\beta}(Q)$ unless it is already at a local minimum. Since $\mathcal{F}_{\beta}(Q)$ is bounded from below by $\mathcal{F}_{\beta}(P^{\text{Gibbs}})$ and each object $u$ is updated infinitely often according to the object visitation schedule, the algorithm converges to a local minimum of the generalized free energy $\mathcal{F}_{\beta}$ within the space of factorial distributions $\mathcal{Q}$.

\end{proof}

\section{Additional Details About Information-Theoretic Acquisition Functions} \label{appendix:IG}

\subsection{Detailed Derivation of Entropy} \label{appendix:mu}

Here we show a detailed derivation of the probability $P(E_{uv})$, which is used for the acquisition function based on entropy in Eq. \ref{eq:entropy}. We have

\begin{equation} \label{eq:euvproof}
\begin{aligned}
    P(E_{uv} = 1) &= \mathbb{E}_{P^{\text{Gibbs}}(\rvy)}[\1_{\{\ervy_u = \ervy_v\}}] \\
    &\approx \mathbb{E}_{Q(\rvy)}[\1_{\{\ervy_u = \ervy_v\}}] \\
    &= \sum_{k^{\prime} \in [K]} Q_{uk^{\prime}} \sum_{k^{\prime\prime} \in [K]} Q_{uk^{\prime\prime}} \1_{\{c_u = c_v\}} \\
    &= \sum_{k \in [K]} Q_{uk}Q_{vk} + \underbrace{\sum_{k^{\prime} \in [K]} \sum_{\substack{k^{\prime\prime} \in [K] \\ k^{\prime\prime} \neq k^{\prime}}} Q_{uk^{\prime}}Q_{vk^{\prime\prime}} \1_{\{c_u = c_v\}}}_{= 0} \\
    &= \sum_{k \in [K]} Q_{uk}Q_{vk}.
\end{aligned}
\end{equation}

One can also show that $P(\ermE_{uv} = -1) \approx \mathbb{E}_{Q(\vc)}[\1_{\{c_u \neq c_v\}}(\vc)]$ which can be simplified to $P(\ermE_{uv} = -1) = \sum_{k,k^{\prime} \in [K]} Q_{uk}Q_{vk^{\prime}}\1_{\{k \neq k^{\prime}\}} = 1 - P(\ermE_{uv} = 1)$.

\subsection{Derivations of EIG} \label{appendix:IGO}

In this section, we present a detailed derivation of the acquisition function defined in Eq. \ref{eq:eigo}. We begin by considering the entropy of the cluster labels,
\begin{equation}
    H(\rvy) = -\sum_{\vc \in \mathcal{C}} P^{\text{Gibbs}}(\rvy = \vc) \log P^{\text{Gibbs}}(\rvy = \vc).
\end{equation}
A mean-field approximation of the Gibbs distribution,
\begin{equation}
    Q(\rvy = \vc) = \prod_{u = 1}^N Q(\ervy_u = c_u),
\end{equation}
assumes independence between the cluster labels $\ervy_u$. With this assumption, the entropy can be expressed as
\begin{equation} \label{eq:igproof}
\begin{aligned} 
    H(\rvy) &\approx -\sum_{\vc \in \mathcal{C}} Q(\rvy = \vc) \log Q(\rvy = \vc) \\
    &= -\sum_{\vc \in \mathcal{C}} \prod^N_{u = 1} Q(\ervy_u = \evc_u) \log \prod^N_{v = 1} Q(\ervy_v = \evc_v) \\
    &= -\sum_{\vc \in \mathcal{C}} \prod^N_{u = 1} Q(\ervy_u = \evc_u) \left(\sum_{v = 1}^N \log Q(\ervy_v = \evc_v)\right) \\
    &= -\sum_{\vc \in \mathcal{C}} \sum_{v = 1}^N \prod^N_{u = 1} Q(\ervy_u = \evc_u) \log Q(\ervy_v = \evc_v) \\
    &= -\sum_{v = 1}^N \sum_{k \in [K]} \sum_{\substack{\vc \in \mathcal{C} \\ c_v = k}} \prod^N_{u = 1}  Q(\ervy_u = \evc_u) \log Q(\ervy_v = \evc_v) \\
    &= -\sum_{v = 1}^N \sum_{k \in [K]} \log Q(\ervy_v = \evc_v)  \sum_{\substack{\vc \in \mathcal{C} \\ c_v = k}}  \prod^N_{u = 1}  Q(\ervy_u = \evc_u) \\
    &= -\sum_{v = 1}^N \sum_{k \in [K]} \log Q(\ervy_v = \evc_v) Q(\ervy_v = \evc_v)  \underbrace{\left(\sum_{\substack{\vc \in \mathcal{C} \\ c_v = k}}  \prod^N_{\substack{u = 1 \\ u \neq v}}  Q(\ervy_u = \evc_u)\right)}_{= 1} \\
    &= -\sum_{v = 1}^N \sum_{k \in [K]} Q(\ervy_v = \evc_v)\log Q(\ervy_v = \evc_v)\\
    &=  \sum_{v = 1}^N H(\ervy_v \mid \mQ).
\end{aligned}
\end{equation}

Assuming independence between the pairwise variables in $\rmE$ given $\mQ$, the joint distribution factorizes as
\begin{equation}
    P(\rmE) = \prod_{(w, l) \in \mathcal{E}} P(\ermE_{wl} \mid \mQ).
\end{equation}
Using an argument analogous to that in Eq.~\ref{eq:igproof}, we obtain
\begin{equation}
    H(\rmE) \approx \sum_{(w, l) \in \mathcal{E}} H(\ermE_{wl} \mid \mQ).
\end{equation}

Furthermore, we approximate the conditional entropies $H(\rvy \mid \ermE_{uv}=e)$ and $H(\rmE \mid \ermE_{uv}=e)$ using a conditional mean-field approximation, denoted by $\mQ^{(S_{uv} = e)}$. This approach decomposes the joint conditional entropy into a sum over the individual entropies, following the same reasoning as in Eq.~\ref{eq:igproof}. Consequently, we arrive at the acquisition function $a^{\text{EIG-O}}$ defined in Eq.~\ref{eq:eigo} in the main paper.

\end{document}